\documentclass[runningheads]{llncs}
\pdfoutput=1 
 
\usepackage{eccv}
\usepackage{pdfpages}

\usepackage{eccvabbrv}

\usepackage{graphicx}
\usepackage{booktabs}

\usepackage[accsupp]{axessibility}  

\usepackage[pagebackref,breaklinks,colorlinks,citecolor=eccvblue]{hyperref}
\PassOptionsToPackage{pagebackref=true}{hyperref}
\usepackage{orcidlink}
\usepackage{textcomp, gensymb}
\usepackage{lipsum}
\usepackage{array} 
\usepackage{booktabs}
\usepackage{placeins}
\usepackage{enumitem}
\usepackage{comment}
\usepackage[nameinlink]{cleveref}

\begin{document}

\title{Helios: An extremely low power event-based gesture recognition for always-on smart eyewear} 

\titlerunning{Helios: An extremely low power event-based
gesture recognition}

\author{Prarthana Bhattacharyya\textsuperscript{*} \hspace{0.2cm} Joshua Mitton\textsuperscript{*} \hspace{0.2cm}  Ryan Page\textsuperscript{*} \hspace{0.2cm} Owen Morgan \\ Ben Menzies \hspace{0.2cm} Gabriel Homewood \hspace{0.2cm} Kemi Jacobs \hspace{0.2cm} Paolo Baesso \\ David Trickett \hspace{0.2cm} Chris Mair \hspace{0.2cm} Taru Muhonen \hspace{0.2cm} \hspace{0.2cm} Rory Clark \\ Louis Berridge \hspace{0.2cm} Richard Vigars \hspace{0.2cm} Iain Wallace
}

\authorrunning{Bhattacharyya, Mitton, Page et al.}

\institute{Ultraleap, Glass Wharf, Bristol, UK \\ 
\email{\{prarthana.bhattacharyya, josh.mitton, ryan.page\}@ultraleap.com}}
\maketitle
\renewcommand{\thefootnote}{}  
\footnotetext{\textsuperscript{*}Authors contributed equally to this paper.}
\vspace{-0.2in}
\begin{abstract}
This paper introduces Helios, the first extremely low-power, real-time, event-based hand gesture recognition system designed for all-day on smart eyewear. As augmented reality (AR) evolves, current smart glasses like the Meta Ray-Bans prioritize visual and wearable comfort at the expense of functionality. Existing human-machine interfaces (HMIs) in these devices, such as capacitive touch and voice controls, present limitations in ergonomics, privacy and power consumption. Helios addresses these challenges by leveraging natural hand interactions for a more intuitive and comfortable user experience. Our system utilizes a extremely low-power and compact 3mm$\times$4mm/20mW event camera to perform natural hand-based gesture recognition for always-on smart eyewear. The camera's output is processed by a convolutional neural network (CNN) running on a NXP Nano UltraLite compute platform, consuming less than 350mW. Helios can recognize seven classes of gestures, including subtle microgestures like swipes and pinches, with 91\% accuracy. We also demonstrate real-time performance across 20 users at a remarkably low latency of 60ms. Our user testing results align with the positive feedback we received during our recent successful demo at \href{https://www.awexr.com/usa-2024}{AWE-USA-2024}. A real-time video demonstration of Helios can be found \href{https://0e84f9dd10852326-tracking-platform-shared-public-assets.s3.eu-west-1.amazonaws.com/IMG_1720.MOV}{at this link}.
\keywords{augmented reality \and microgestures \and natural interactions \and machine learning \and smart eyewear \and low-power \and real-time}
\end{abstract}

\section{Introduction}
\label{sec:intro}
\begin{figure}[t]
    \centering
    \includegraphics[width=0.5\textwidth]{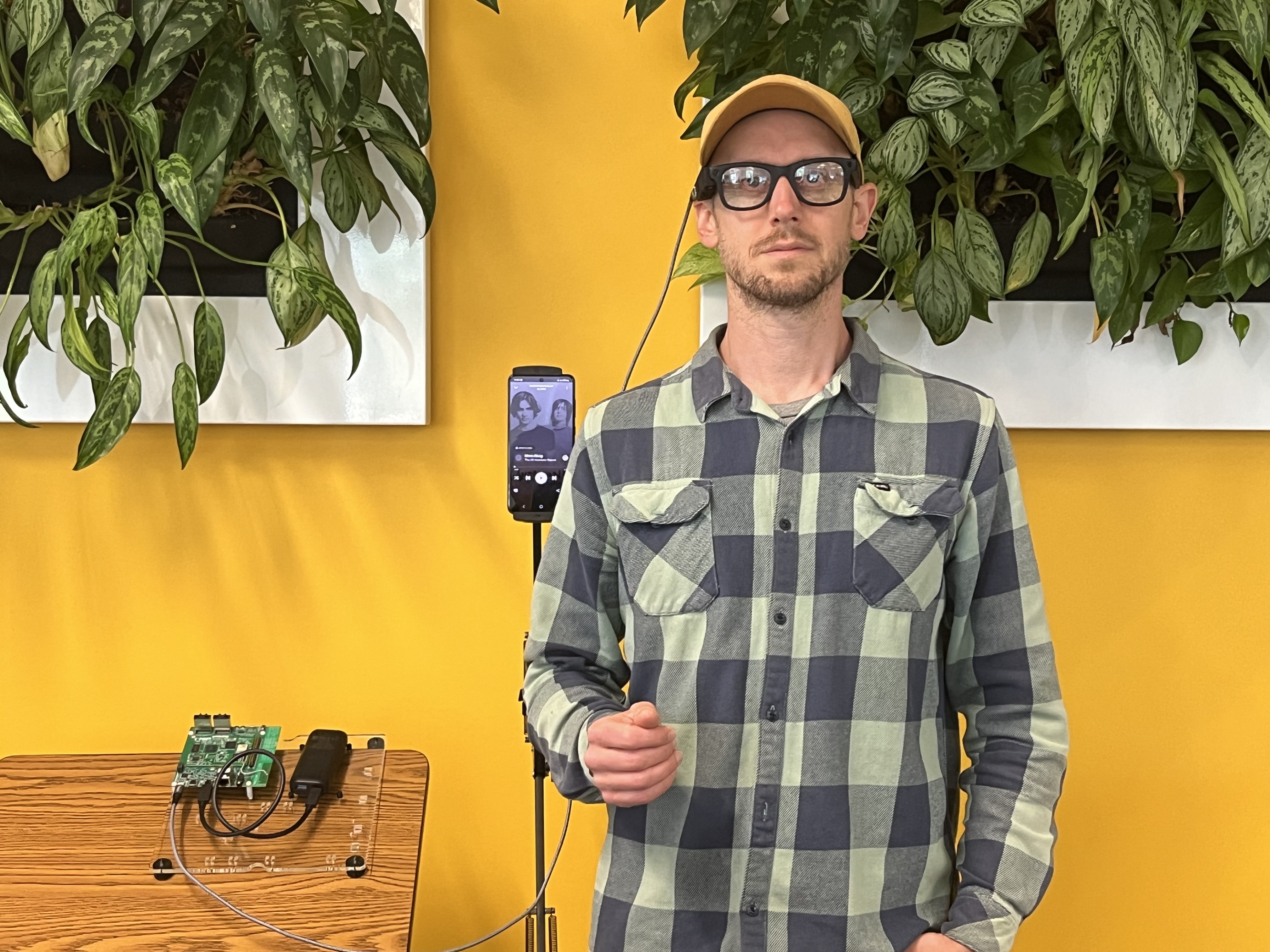}
    \caption{Picture of using the setup: a user wearing smart glasses with event sensors, and controlling Spotify on the mobile device via natural hand gestures, instead of capacitive touch or voice.}
    \label{fig:finalsetup}
\vspace{-0.2in}
\end{figure}
The most successful smart glasses, for example \href{https://about.fb.com/news/2023/09/new-ray-ban-meta-smart-glasses/}{Meta Ray-Bans}, are prioritising what technology can be integrated to the familiar and socially acceptable form factor of glasses, rather than compromising this in favour of spatial computing features which have yet to reach maturity. The use cases range from open ear audio, bringing convenience from a smartphone to the user, to enabling a modality for interacting with AI assistants. For practical use such a device needs to be always on and last all day on a single charge, such that a user can interact with the device when required. This should not come at the expensive of wearable comfort and as such the device weight has to be considered. As a result it is important to develop a low power on device solution that has high functionality.
\par The Human Machine Interface (HMI) is dominated by capacitive touch on the device and voice controls. Capacitive touch interfaces require direct contact with the device's surface, which can lead to discomfort, especially during prolonged use. Touchscreens can be less responsive and cumbersome to use in certain conditions: such as when users are wearing gloves or when the screen is wet or dirty. Additionally, the placement of head worn glasses is not ergonomic to interact with and can be obstructed by headwear and long hair. On the other hand, voice is very expressive. However, voice suffers from an element of social discomfort, which is especially true when considering wake words. On top of that voice controls are often switched off to limit power consumption.
\par In contrast, natural hand interactions offer a more ergonomic and comfortable way to interact with devices. Gestural interfaces allow users to perform actions with simple hand movements, which can be more intuitive. Gesture-based controls can provide a seamless and immersive user experience, making interaction with technology more accessible and enjoyable. Further, gesture-based interactions can be more discrete and avoid the social discomfort associated to voice-based approaches.
\par Gesture recognition systems can be used to interpret hand movements, and translating them into specific commands for media control. For instance, swiping, can be mapped to actions such as navigating through a media library. Due to their resemblance to smartphone interactions, these gestures are intuitive and easy to learn. By linking natural hand movements to media control through gesture recognition, it opens up new possibilities for innovation in how we interact with our devices.
\par Recognizing microgestures, characterized by small motions such as swipes or pinches, presents a significant challenge due to their subtlety and the difficulty in distinguishing them from hand tracking jitter or noise. Additionally, user variation complicates recognition with simple heuristics. To address this, we apply a machine learning (ML) approach using convolutional neural networks (CNN) across seven gesture classes. Real-time hand-gesture recognition is crucial for applications in augmented reality, which requires low latency for smooth interaction typically within 100 ms. Conventional cameras often struggle with motion artifacts and challenging lighting conditions, leading to increased power consumption. We leverage event-cameras to mitigate these issues, providing an efficient solution for real-time gesture recognition. Unlike traditional frame-based cameras that capture redundant data at fixed intervals, event cameras only record changes in brightness, resulting in a sparse and efficient data stream. This translates to lower latency and low power requirements.
\par This paper describes Helios: a low-power, real-time, event-based hand gesture recognition system for smart eyewear. Events from a 3mm$\times$4mm/20mW event camera are passed to a convolutional neural network running on NXP Nano UltraLite which achieves significant power and speed advantages. Helios consumes less than 350mW and has a latency of 60ms from the start of the gesture. In contrast to Meta's \cite{STMG}, we do not use frame cameras, and our gesture recognition does not work on top of hand skeletal tracking. Helios's novelty is in being able to go from event streams to gesture recognition directly. The closest to our work is \cite{GesturewithEvents_2017} which also uses events for gesture recognition. They however use the \href{https://inivation.com/wp-content/uploads/2019/08/DVS128.pdf}{DVS-128} camera which is 40mm$\times$60mm and unsuitable for use with smart eyerwear. A key feature of event-based gesture recognition to work for smart eyewear is robustness to false positives with ego-motion. We expect the user to be walking around while using the smart eyewear. Helios models adversarial classes along with relevant microgestures so that our gesture recognition system does not get triggered by random noise due to human movement. This in contrast to \cite{aliminati2024sevd, GesturewithEvents_2017, STMG} where this problem is not actively addressed. Our complete setup is shown in \Cref{fig:finalsetup}. \\ 
\textbf{Contributions:} To the best of our knowledge, this is the first real-time gesture recognition system that enables the use of natural hand interactions with smart eyewear. Helios's gesture recognition system is implemented on hardware that operates on event streams in real-time and enables a user to interact with smart glasses efficiently, comfortably and robustly without capacitive touch or voice. We also present the results of a user study with Helios, that closely matches the experience of our recent successful demo at \href{https://www.awexr.com/usa-2024}{AWE-USA-2024}. 

\section{Related Work}
\subsubsection{Event Camera Background:} Event sensors 
promise unprecedented temporal resolution for extremely low power, as low as 3~mW. This contrasts to frame based sensors that require trading temporal resolution with power and typically consume between 35-200~mW depending on frame rate, for example \href{https://www.st.com/en/imaging-and-photonics-solutions/vd55g1.html}{ST's VD55G1} and \href{https://www.ovt.com/products/og01a1b/}{Omnivison's OG01A1B}. Another key advantage is that event sensors have inherently high dynamic range, allowing systems to work in bright sunlight and moonlight. Again, this requires some tradeoffs to achieve in frame based devices. These two factors, ultra low power and high dynamic range, are fundamental for wearable devices  making event sensors an attractive technology. The underpinning technology is biologically inspired and measures per pixel changes in the brightness outputting an event when the change is above a certain contrast threshold~\cite{EventsSurveyPaper}. This results in a sparse data stream of the form $e_{k}:=(x_{k}, y_{k}, p_{k}, t_{k})$ where $x_{k}, y_{k}$ are the $(x, y)$ pixel coordinates respectively, $p_{k}$ is the polarity defined as 1 for positive changes and 0 for negative changes and $t_{k}$ is the time the event occurred, typically in $\mu s$ for event $k$. This is a very different data format compared with a frame based camera which outputs brightness for every pixel periodically. Machine vision has been dominated by image processing architectures that work effectively on frames, however those systems can not meet the needs of wearable technology and this is what motivates research into event based vision. The following sections go into the challenges this presents the machine learning community and outlines the current landscape. 
\vspace{-0.3cm}
\subsubsection{Event Representations}
Two of the most common ways to represent event data for machine learning applications are event volumes and time surfaces \cite{EventsSurveyPaper}. Time surfaces \cite{EventHandsICCV, Prophesee_paper_detection_AVs, EventPoint} retain the timestamp of the most recent event for each pixel and polarity, creating a 2D map where each pixel stores a single time value. This representation compresses information effectively but loses effectiveness on textured scenes where pixels spike frequently. Event volumes \cite{EventstoVideo_19, Prophesee_paper_detection_AVs}, on the other hand, are 3D histograms of events, preserving better temporal information but potentially discarding polarity information through voxel accumulation.
\vspace{-0.3cm}
\subsubsection{Event ML Architectures}
Events from event cameras are asynchronous and spatially sparse, whereas images are synchronous and dense. This disparity means that traditional frame-based vision algorithms designed for image sequences are not directly applicable to event data. Current machine learning methods for classification and detection typically aggregate events into a time surface or an event volume and process them using standard vision pipelines like Convolutional Neural Networks (CNN) or recursive architectures such as Conv-LSTM \cite{millerdurai2024eventego3d, liang2024towards, gao2024sd2event, chen2024segment, kong2024openess, aliminati2024sevd, EventstoVideo_19}. These setups, while simple, often involve wasteful computations and can be computationally expensive. Initial methods for event-based machine learning focused on standard vision architectures like CNN and Conv-LSTM, but more specialized architectures inspired by LIDAR object detection have emerged. Researchers have explored sparse convolutions \cite{liu2015sparse}, which compute convolutions only at active sites with non-zero feature vectors, reducing redundant computations \cite{peng2024scene, yu2024eventps, zhang2024co}. Leveraging temporal sparsity involves retaining previous activations and applying recursive sparse update rules \cite{ren2024simple, SparseEventGraphGNN}. Unlike time surfaces and voxel grids, graph representation-based approaches \cite{SparseEventGraphGNN, GNNsforEvents} preserve the compactness and asynchronicity of event streams, with Graph Neural Networks (GNNs) efficiently handling irregular and sparse data structures. In summary, while traditional vision architectures have been adapted for event data, more specialized and efficient methods leveraging the unique properties of event streams continue to be developed, showing promise for a variety of applications.

\section{Machine Vision System}
\label{hardware}
\sloppy

\begin{figure}[t]
    \centering
    \includegraphics[angle=90, width=0.7\textwidth]{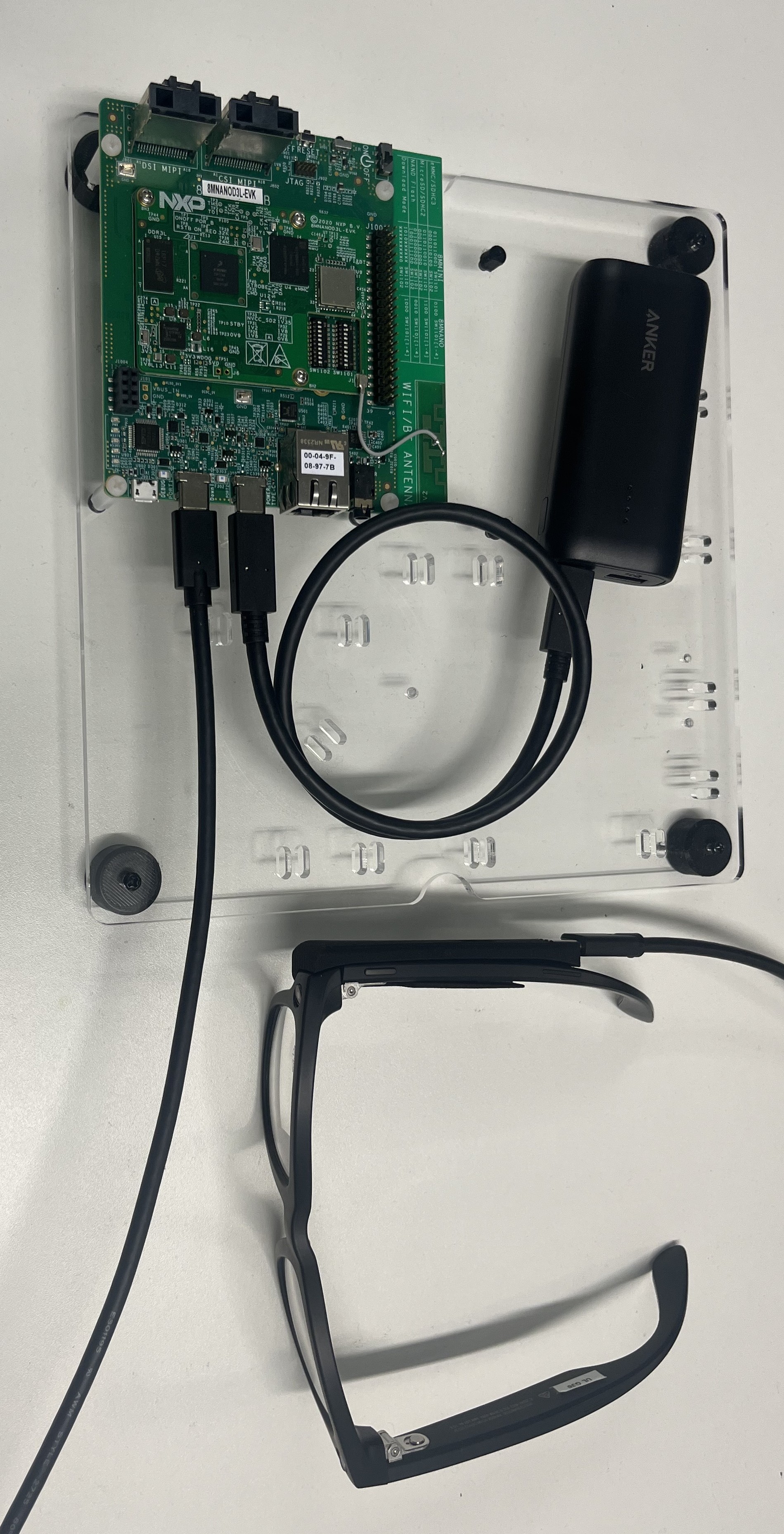}
    \caption{Our hardware setup consisting of Prophesee GenX320 event camera fitted onto Meta Ray-Ban smart glasses connected to a NXP iMX 8M Nano UltraLite for processing events}
    \label{fig:cameraboardglasses}
    \vspace{-0.2in}
\end{figure}

The machine vision system was designed and optimised to enable the Human Machine Interface (HMI) discussed in \Cref{ux-research} to work with a glasses mounted mono event camera setup. The complexity of doing a full integration into a pair of smart frames or glasses led to an architecture consisting of two parts connected over USB2.0. The first was an event camera USB peripheral consisting of the Prophesee GenX320 sensor and the Cypress CX3 MIPI CSI2 - USB bridge, the event stream was then transmitted over USB2.0 to the second part, a NXP iMX 8M Nano UltraLite compute platform. This arrangement is shown in \Cref{fig:cameraboardglasses}. The compute platform is designed for power-efficient devices using ARM Cortex-A53 cores which aligns with current smart glass optimised mobile processors. The following sections go into details for each of these components.
\vspace{-0.2cm}
\subsection{Event Camera Peripheral}

The event camera peripheral was a custom built device with two major parts, the event camera module and the camera controller chip. The event camera module used was the Compact Optical Module (CM2) from  \href{https://www.prophesee.ai/event-based-sensor-genx320/}{Prophesee}. This has package dimensions of $8 \times 8 \times 5~mm$, a field of view of 84$^{o}$ for both horizontal and vertical directions and 104$^{o}$ for the diagonal, and features a GenX320 sensor. The GenX320 is an array of $320\times320$ $6.3~\mu m$ event pixels that features a number of readout and event signal processing modes. This includes a low power monitor mode which consumes 2.9mW, a parallel readout mode that depending on event rate consumes between $3 - 4$~mW, and a MIPI CSI2 streaming mode that consumes 22.8~mW. The latter is used in the studies presented here. The low light cutoff is 0.05~lux, enabling nearly full coverage of day and night time illuminance. This module is then connected over a flexible printed circuit (FPC) to the main PCB which has a CX3 MIPI CSI2 to USB camera controller chip. This converts the CSI2 data stream to USB traffic that is sent to a host device. 
\begin{figure}[t]
    \centering
    \includegraphics[width=0.5\textwidth]{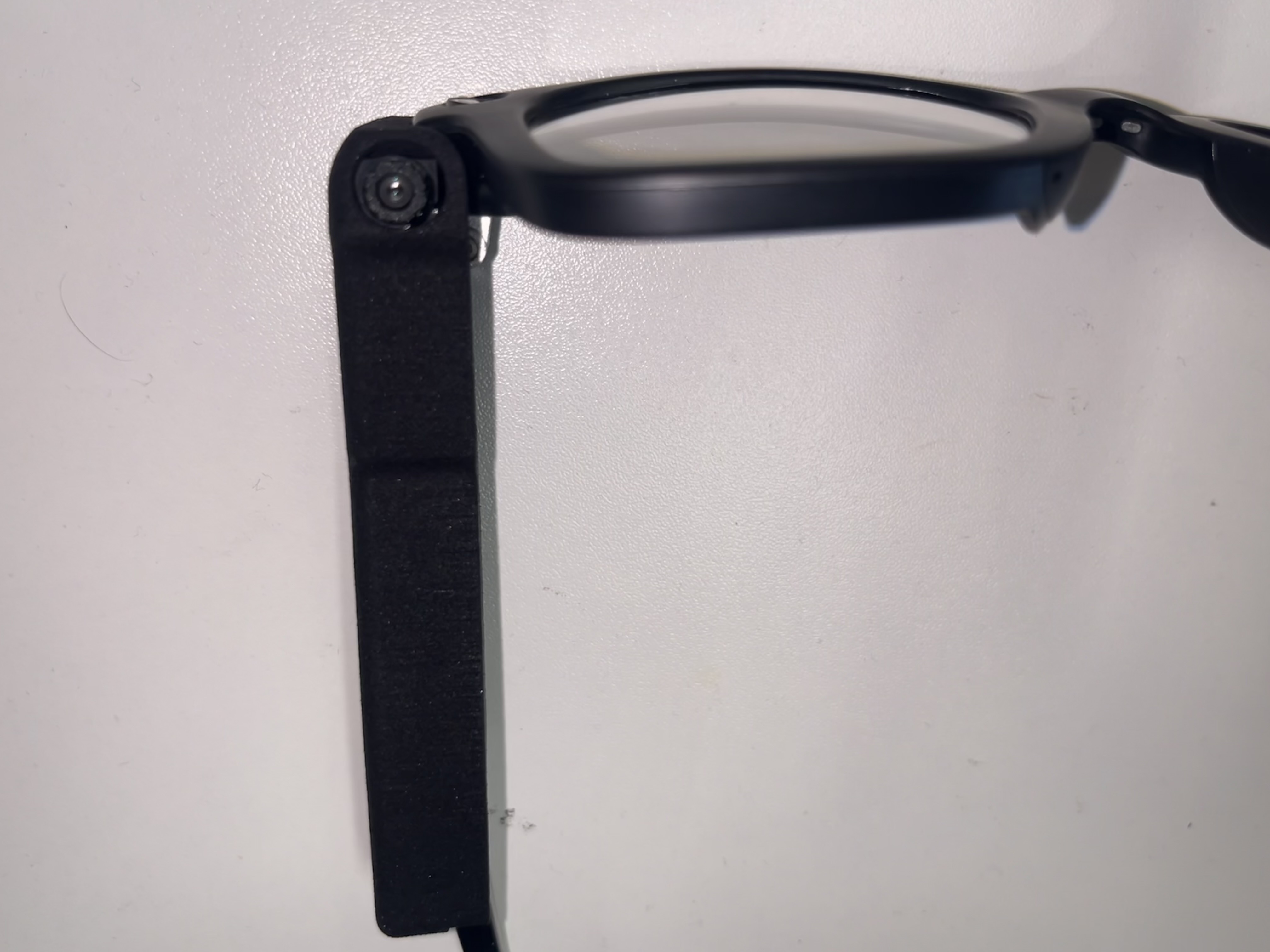}
    \caption{In this image the camera module can be seen in the housing that is attached to the Meta Ray-Ban glasses. The camera is pointing down with a small tilt away from the body of the user}
    \label{fig:smartglasses}
    \vspace{-0.2in}
\end{figure}

\subsection{Compute Platform}
The event stream is consumed by our tracking service running on a development platform that contains the i.MX 8M Nano UltraLite processor from NXP, along with 1~GB of RAM. The processor comprises 4x Arm Cortex A-53 cores running at 1.2-1.4~GHz using the Armv8-A instruction set. These cores are optimised for efficiency making them attractive for mobile devices. The operating system was embedded linux with kernel version 6.1.22 distributed with the development platform and built from the i.MX board support using Yocto. The board can be seen in \Cref{fig:cameraboardglasses}, where the USB connection to the peripheral can be seen, along with the battery pack also connected over USB.
\subsection{System assembly}
The event camera module and main PCB were assembled into a enclosure designed to fit onto the Meta Ray-Ban glasses. The camera is oriented such that the camera's optical axis is pointing downward with a small $20^{o}$ rotation around the camera's $X$-axis (user is facing the Z axis), based on the OpenCV coordinate system, to accommodate an interaction volume further from the body. Our smartglasses setup is illustrated in \Cref{fig:smartglasses}. Finally, the device is connected to the compute platform detailed above and is compatible with any applications built using \href{https://github.com/prophesee-ai/openeb}{OpenEB} and the \href{https://docs.prophesee.ai/stable/index.html}{MetaVision SDK} from Prophesee.

\section{UX Research}
\label{ux-research}
The user experience we are looking to develop enables smart glasses to be controlled with simple intuitive gestures.

\textit{What could the basic gesture controls look like?} This could include but is not limited to basic controls using the thumb’s retroposition/opposition movements to create left/right navigation and flexion/extension to create up/down navigation and thumb’s palmer abduction/adduction in combination with finger flexion/extension to create pinch for select. On top of this, the dynamics of the motion can encode a variation on the action. In the example of a media control, a right swipe can skip track, while a double swipe can encode move forward 30 seconds. These motions can be identified through spatiotemporal pattern recognition. In the instance where an event camera is used the temporal information can be presented in different ways. An example is embedding time in the spatial domain. This is most advantageous when interaction motions are at a higher frequency to the background motion so time windows and pixel can be tuned to extract hand motion and reject background motion. To decide the basic control, internal testing was conducted with 9 internal participants.  The initial version of interactions are focused on smart glass media controls, enabling users to change track, volume, or play/pause. The four stages of UX research undertaken are:
\begin{figure}[t]
    \centering
    \includegraphics[width=0.7\textwidth]{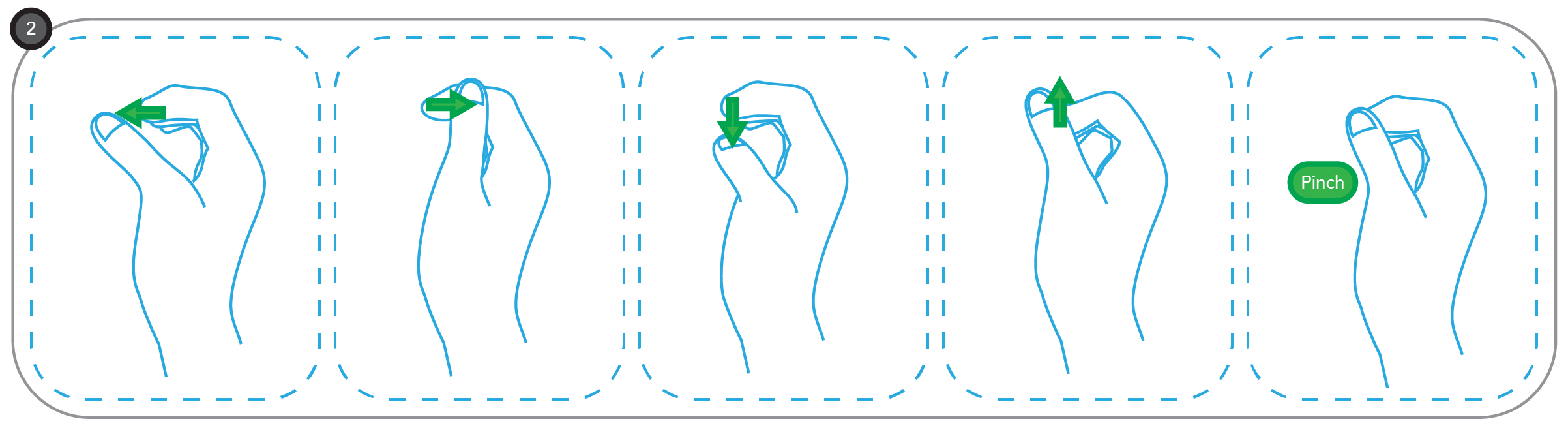}
    \caption{Our chosen microgestures for natural hand interactions with smart eyewear: thumb swipes and pinch.}
    \label{fig:microgestures}
    \vspace{-0.2in}
\end{figure}

\begin{enumerate}
    \item Determine the potential hand interactions for media control.
    \item After using each of the interactions, participants were asked to score the interaction between 1-5 for different factors.
    \item Determine which gestures can lead to a good interaction score for users, as well as generate a strong signal for event camera signal processing.
    \item After deciding the gestures, find the time taken by different users to complete these gestures.
\end{enumerate}
The gestures we chose for Helios after steps 1,2 and 3 are a combination of pinch and thumb flick/swipes as illustrated in \Cref{fig:microgestures}. People liked how small the gesture could be, meaning lower effort and less noticeability in public. These gestures require no hand movement, only finger and thumb, making these the smallest interactions tested. The interactions were scored as the most socially acceptable with users most likely to feel comfortable performing these in public. 
By visual inspection, we verified that this set of gestures produces a signal sufficient for a machine learning model to learn from. Therefore, this is also an encouraging choice from a machine learning perspective. 

For the final step 4, we record the users doing the gestures and study how much time it takes to complete the gestures on average. 
This information is used to determine the aggregation window of event representations for machine learning based gesture control, as described in more detail in \Cref{sec:representation}. The period of time chosen for the event aggregation into an event representation is important as the event representation captures the dynamics of action. We experimented with this aggregation period to ensure that it matches the observed frequency of motion. A machine learning model being able to learn from such a representation makes intuitive sense given that our chosen gestures are mostly singular in direction, occur within a given time-period, and time surfaces encode temporal information within the representation. 
\section{Methodology}
\label{sec:method}
Our proposed gesture recognition algorithm has two major components. First, the event sensor captures an event stream. This is then converted into an image-like representation. We specifically use time surfaces \cite{EventsSurveyPaper} as our event representation. Second, the event representation is presented as input features to a machine learning model trained offline using GPU acceleration. This model outputs a classification prediction for each time surface. The entire workflow is described in this section.
\subsection{Event Representation}
\label{sec:representation}
Time surfaces (TS) retain the time stamp of the most recent event for each pixel and polarity. It is a 2D map where each pixel stores a single time value. The advantage of this representation, compared to other grid-like representations, is that TS are highly efficient as a result of only keeping one timestamp per pixel. 
In this paper, we use polarity separated time surface as our event representation. The straightforward way to convert the event stream to a 2D representation is to accumulate and collapse all events in a time interval, which leads to the loss of temporal resolution within the interval. We build on the representation proposed in Event Hands \cite{EventHandsICCV} called Locally-Normalised Event Surfaces (LNES), which encodes all events within a fixed time window as an image $\mathcal{I} \in \mathbb{R}^{\{H, W, 2\}}$. \\
Using separate channels for positive and negative events preserves the polarities and reduces the number of overridden events. To build the TS representation we consider the $k$-th time window in an event stream of size $L$. This window $\mathcal{I}_k$ is created by first initializing it with zeros and collecting events that have timestamps $t_i$ within this window by iterating through from the oldest to the newest event. $t_L$  denotes the timestamp corresponding to the maximum window size $L$. We use an exponential TS instead of a linear TS as proposed in \cite{EventHandsICCV} for an enhanced dynamic range, as given in \Cref{eq:1}.
\begin{equation}
\label{eq:1}
\mathcal{I}_{k} = e^{-{\dfrac{t_L - t_i}{L}}} 
\end{equation}
The window size $L$ is important to choose properly: if it is too long, the TS can be overridden when a new event with the same polarity occurs at the same pixel location; if it is too short, it can be computationally expensive. 
For our experiments, we experimented with the ideal window size and used a fixed time length window for all microgestures. 
\subsection{Machine Learning Model}
\label{sec:ml}
In this section we provide details on our machine learning model for the task of microgesture detection.
\vspace{-0.2in}
\subsubsection{Model Interface}
The input to our model is a TS as detailed in \Cref{sec:representation}. The model predicts probabilities of the input TS containing each of the seven output classes, where the seven output classes contain both microgestures and some adversarial classes included to improve microgesture classification when testing on a real device.
\vspace{-0.2in}
\subsubsection{Architecture Details}
\begin{figure}[t]
  \centering
  \includegraphics[width=0.8\linewidth]{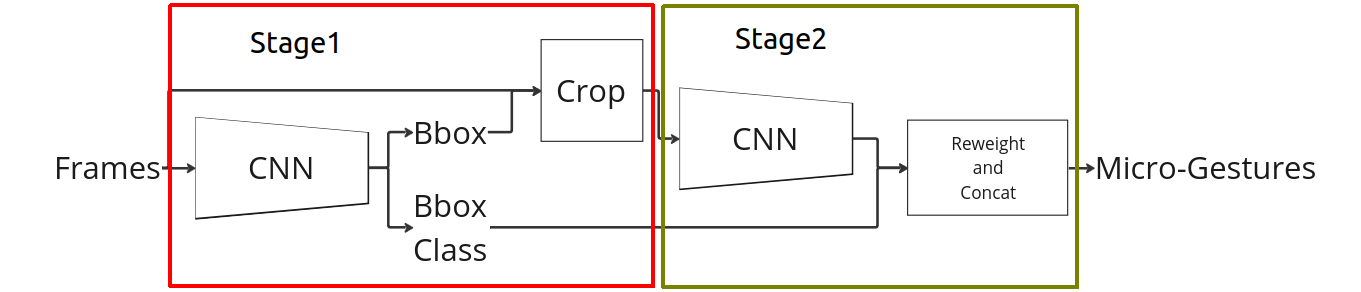}
  \caption{High level block diagram of the gesture detection model architecture}
  \label{fig:model}
\end{figure}

Our model, as shown in \Cref{fig:model}, is a two stage model. The first stage takes TS as an input and predicts a bounding box of the hand in the input and a class prediction of whether there is a hand or not in the bounding box. The TS is then cropped to the bounding box. The second stage takes as input the cropped TS and the bounding box class prediction. It then predicts the probability of the cropped TS containing each of the microgestures. The final model output is then calculated by multiplying the microgesture probabilities with the probability that there was a hand in the TS and concatenating this with the probability of there not being a hand. 
\vspace{-0.2in}
\subsubsection{Loss Function}
When training our model we use two loss functions. The first is a bounding box loss $\mathcal{L}_{\text{bbox}}$, which computes the mean squared error between predicted and true bounding boxes when the true bounding box has a hand in it and provides no signal when the true bounding box does not have a hand. The second loss function $\mathcal{L}_{\text{gesture}}$ is a sparse categorical cross entropy loss between the predicted and true microgestures classes. We found that $\mathcal{L}_{\text{bbox}}$ is required for two reasons. First, the signal from $\mathcal{L}_{\text{gesture}}$ is not strong enough to guide the model to predict bounding boxes that produce TS crops that have the hand centered. Secondly, without $\mathcal{L}_{\text{bbox}}$, the model first learns to distinguish the no hand class from the others and then begins to start learning to distinguish the other microgesture classes, which causes a drop in performance of the final model suggesting it got stuck in a local minimum.
\vspace{-0.2in}
\subsubsection{Model Training}
We trained the network for 35 epochs with a batch size of 1024, using the Adam optimizer with an initial learning rate of 0.0005. We utilize a learning rate scheduler, which linearly decays the learning rate after an initial period where the learning rate is held fixed. We also apply dropout with 0.2 drop rate on the dense layers.
\vspace{-0.2in}
\subsubsection{Model Inference}
At inference time the model requires an aggregated TS as the input. To avoid sampling issues where the microgesture occurred during a period that intersects two models inputs, we create an $L$ window-sized aggregated TS after every 80ms period. The intuition for this is that the model will be unsure on predictions where non-complete microgestures have occurred in the input frame, which will be assigned low probabilities by the model. On the other hand, the predicted probability will be higher when the microgesture occurred within the aggregation window. We overcome the issue of the model predicting a microgesture with low confidence by introducing a \textit{threshold} on softmax probabilities, at which a microgesture is deemed to have occurred. 

\subsection{Dataset collection for training ML gesture recognition system}
Due to the lack of event stream datasets for hand pose estimation or microgesture detection and the difficulty of obtaining accurate ground-truth annotations on real data, we build a highly efficient event stream simulator to generate a large-scale synthetic events dataset with annotations. Our seven classes of interest include: double-pinch (dp), left-swipe (sl), right-swipe (sr), single/moving pinch (mp), resting hands (r), no-hands (nh) and random hand motion (hu). We generated examples of simulated event data per microgesture class for training and validation. The exact number of samples for training and validation sets per gesture class is given in \Cref{fig:combined_confusion_matrix}. 
For our testing setup, we use live user testing as described in \Cref{usertesting}. 
\par
\begin{figure}[t]
    \centering
    \begin{subfigure}[b]{0.45\textwidth}
        \centering
        \includegraphics[width=\textwidth]{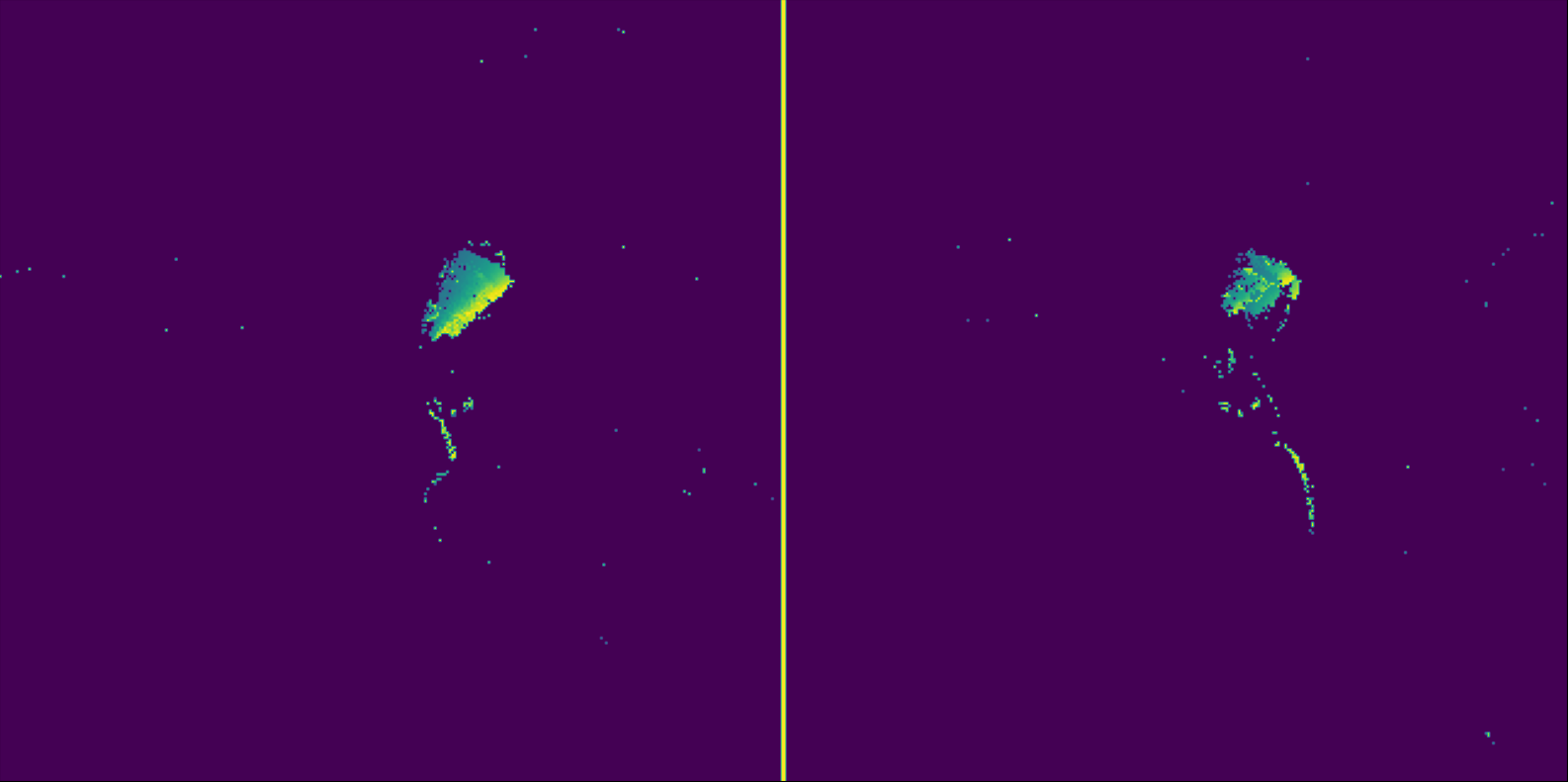}
        \caption{Simulated time surface image for a right swipe}
        \label{fig:ts_subimage1}
    \end{subfigure}
    \hspace{0.05\textwidth}
    \begin{subfigure}[b]{0.45\textwidth}
        \centering
        \includegraphics[width=\textwidth]{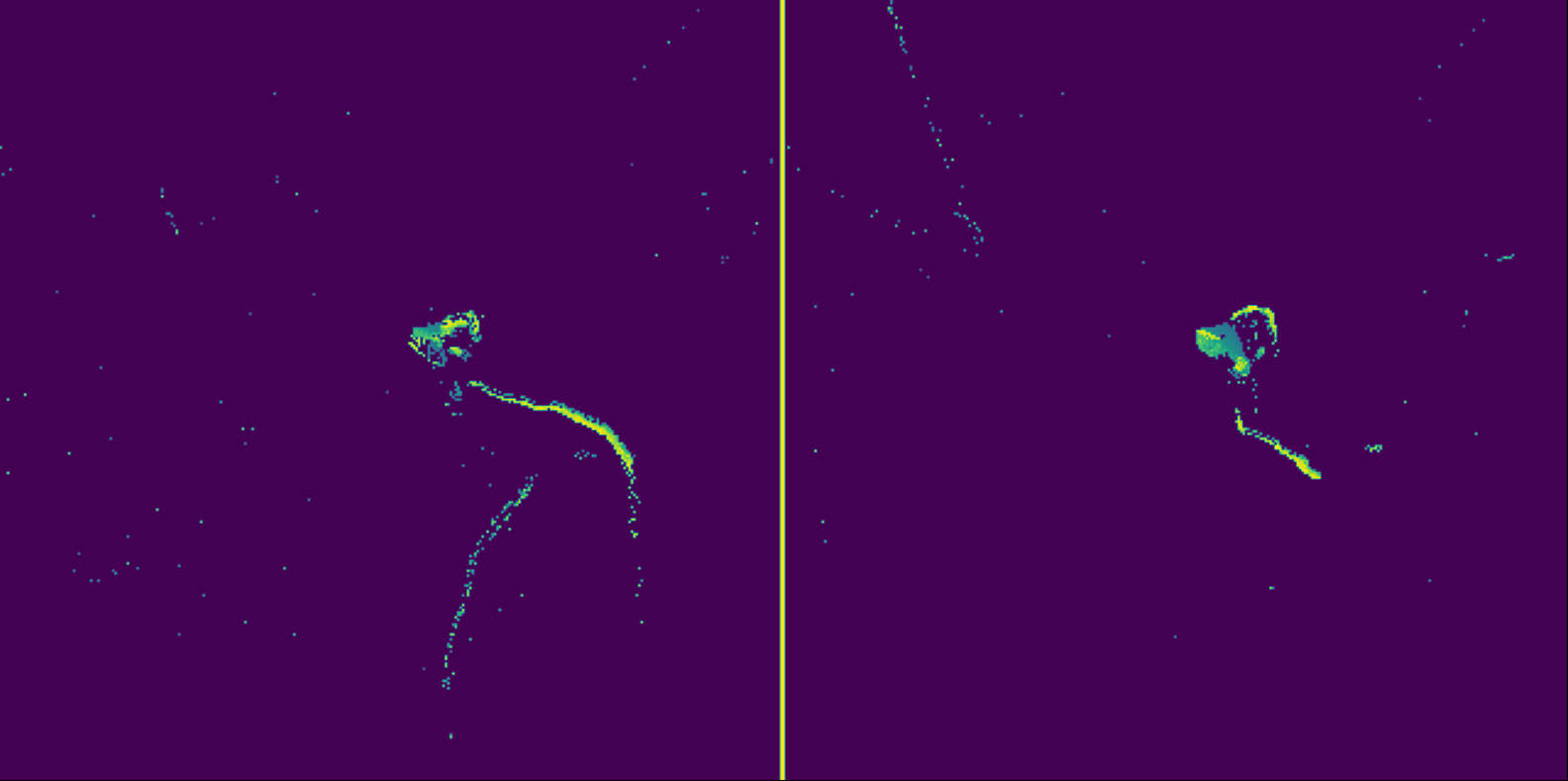}
        \caption{Simulated time surface image for a left swipe}
        \label{fig:ts_subimage2}
    \end{subfigure}
    \caption{Time surface training images for our simulated gestures}
    \label{fig:combined_microgestures}
\vspace{-0.2in}
\end{figure}

 We build on the ESIM simulator \cite{ESIM} to generate training data for the gesture classes. 
Simulating an event camera requires access to a continuous representation of the visual signal at every pixel, which is not feasible in practice. To address this, previous works have proposed sampling the visual signal synchronously at a very high frame rate and using linear interpolation between samples to create a piecewise linear approximation of the continuous visual signal. We adopt this approach for simulating events, by sampling the visual signal from recorded videos of the microgesture being performed. An example of TS generated from simulated data for left swipe and right swipe microgestures is given in \Cref{fig:combined_microgestures}.

\section{Experimental Results}
\label{sec:experiments}
\subsection{Training and Validation Results}
In this section, we present the results of our experiments, focusing on the training and validation accuracy, the behavior of the loss functions, and the analysis of the validation confusion matrix. The train and validation datasets are a 90\% / 10\% split at the sequence from the simulated dataset.
\vspace{-0.2in}

\subsubsection{Training and Validation Accuracy}
\begin{figure}[t]
    \centering
    \begin{subfigure}[t]{0.45\textwidth}
        \centering
        \includegraphics[width=\textwidth]{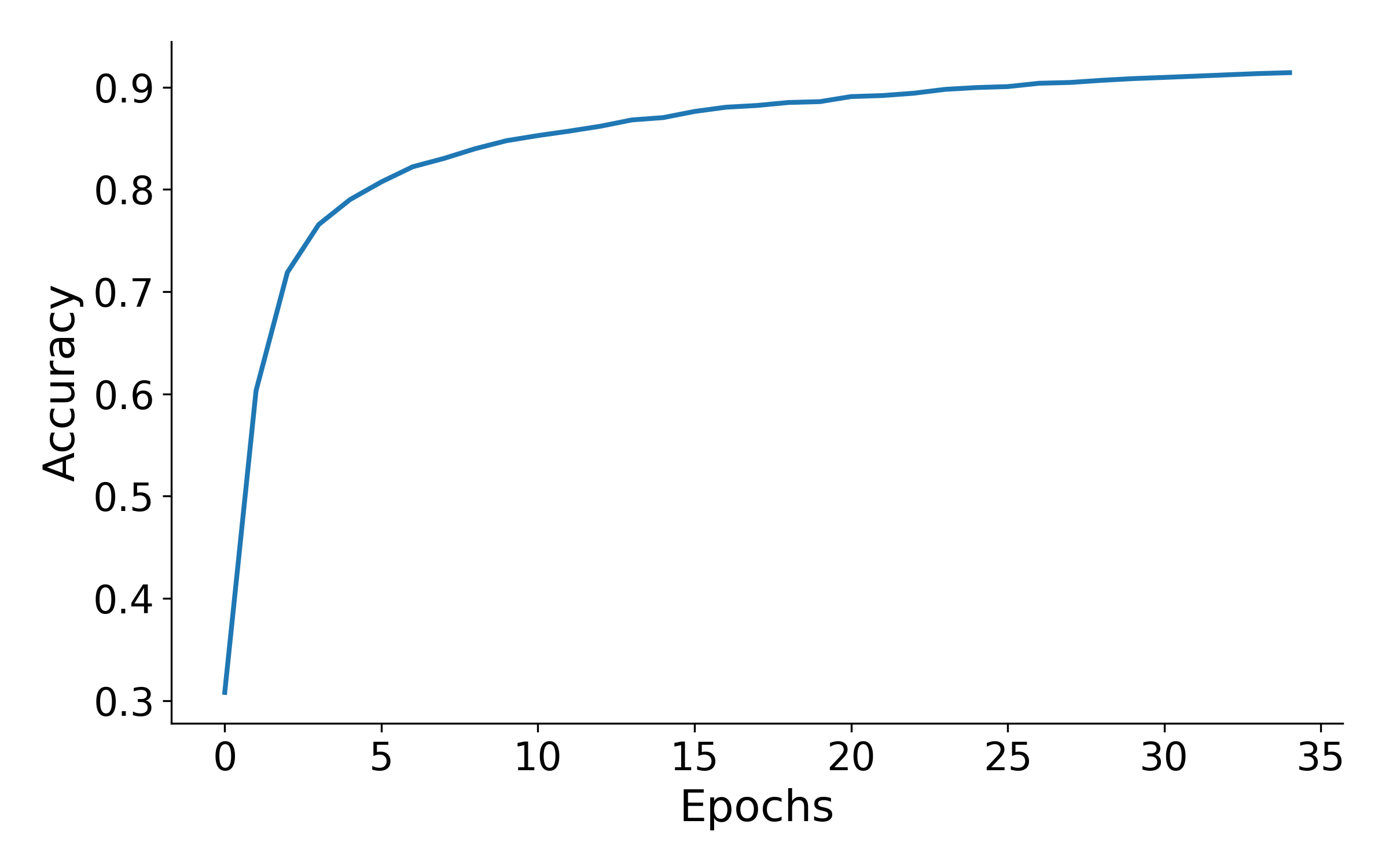}
        \caption{Gesture accuracy for training data}
        \label{fig:accsubimage1}
    \end{subfigure}
    \hspace{0.05\textwidth}
    \begin{subfigure}[b]{0.45\textwidth}
        \centering
        \includegraphics[width=\textwidth]{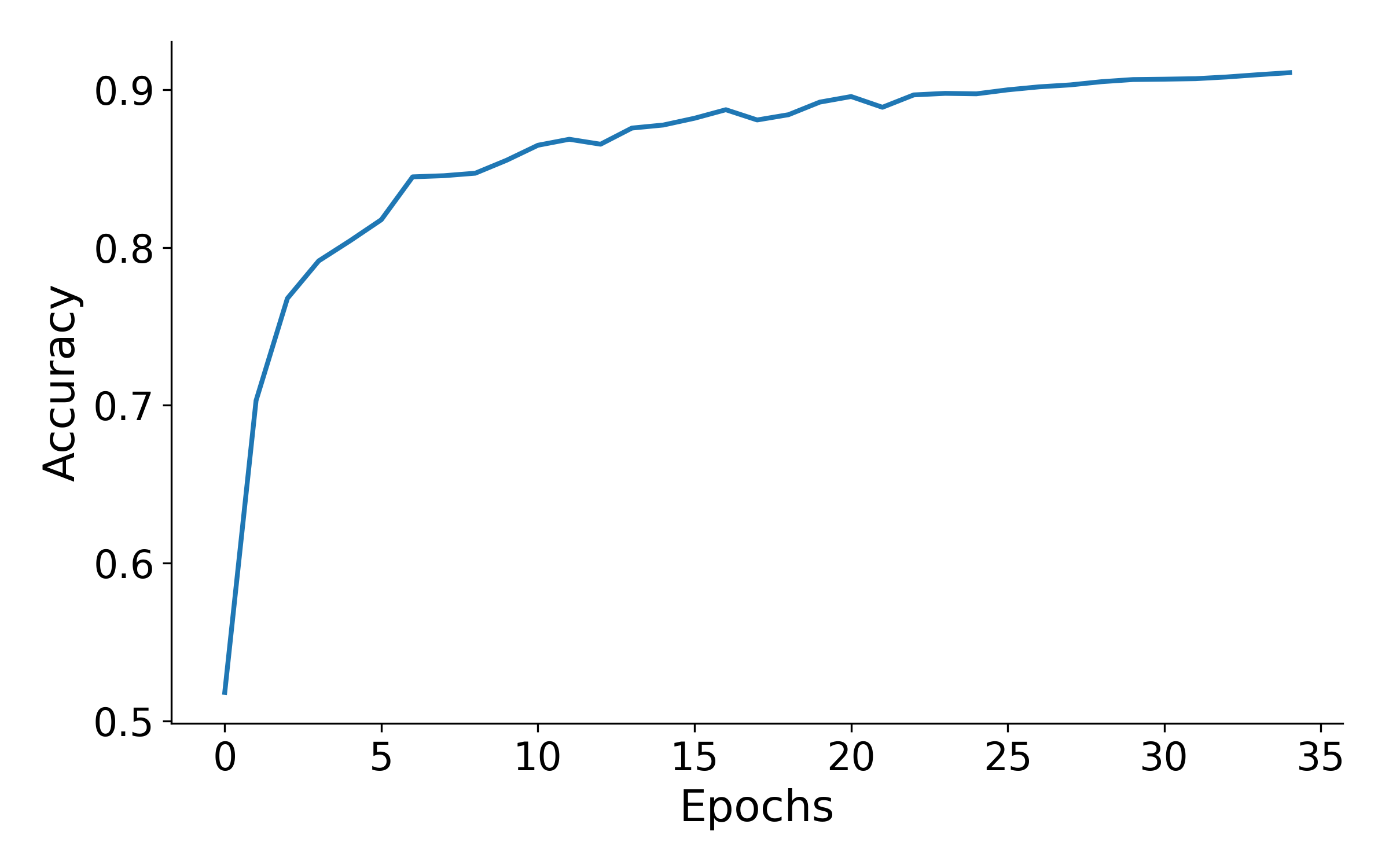}
        \caption{Gesture accuracy for validation data}
        \label{fig:accsubimage2}
    \end{subfigure}
    \caption{Accuracy for the gesture recognition system for training and validation}
    \label{fig:combined_train_val}
    \vspace{-0.2in}
\end{figure}

The model was trained and evaluated on the dataset, achieving a training accuracy of 91.4\% and a validation accuracy of 91\%, as shown in \Cref{fig:combined_train_val}. These results indicate a strong performance of the model on both seen and unseen data, suggesting that the model generalizes well from the training data to the validation set.
\vspace{-0.2in}
\subsubsection{Loss Functions}
\begin{figure}[t]
    \centering
    \begin{subfigure}[b]{0.45\textwidth}
        \centering
        \includegraphics[width=\textwidth]{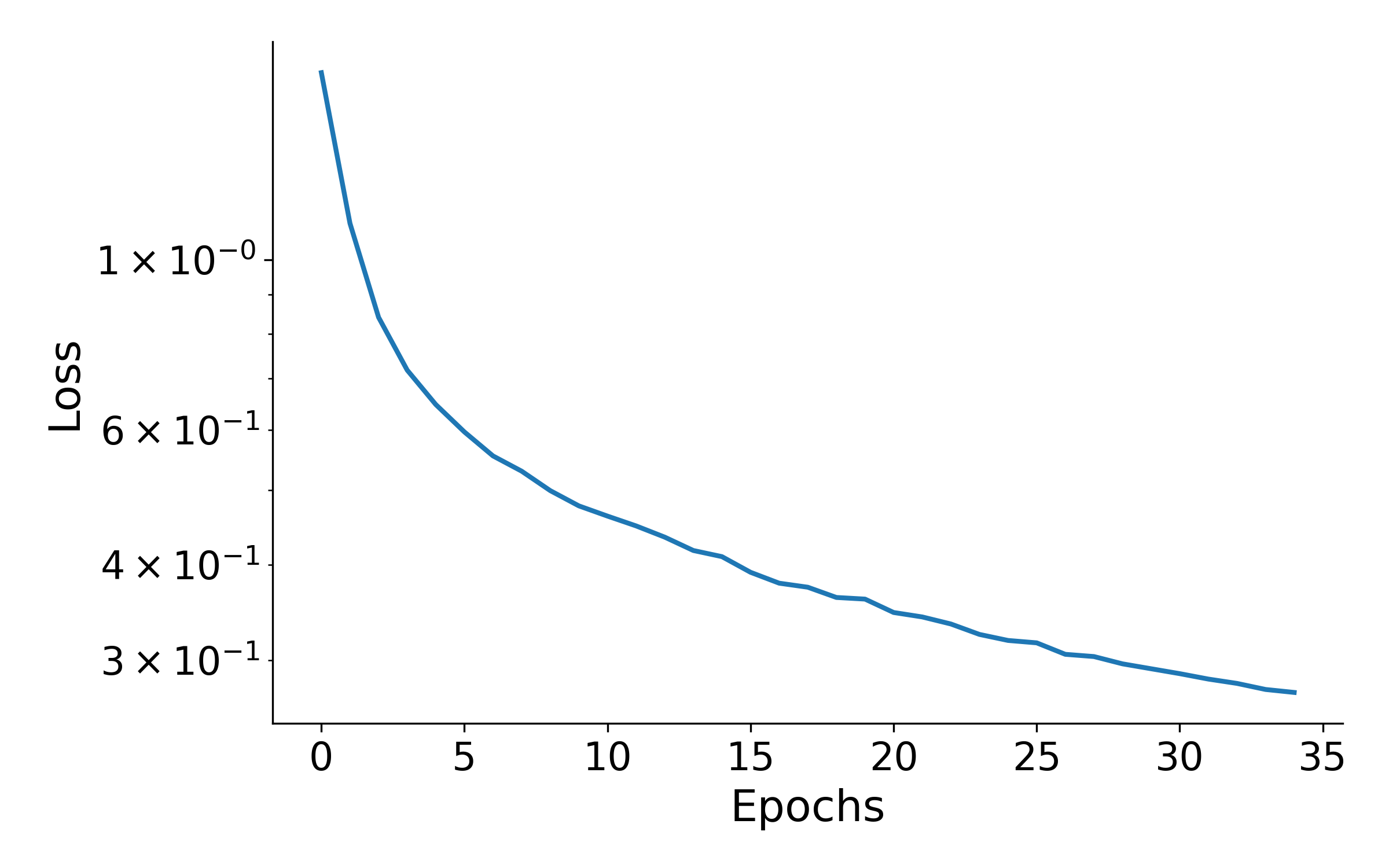}
        \caption{Gesture loss}
        \label{fig:losssubimage1}
    \end{subfigure}
    \hspace{0.05\textwidth}
    \begin{subfigure}[b]{0.45\textwidth}
        \centering
        \includegraphics[width=\textwidth]{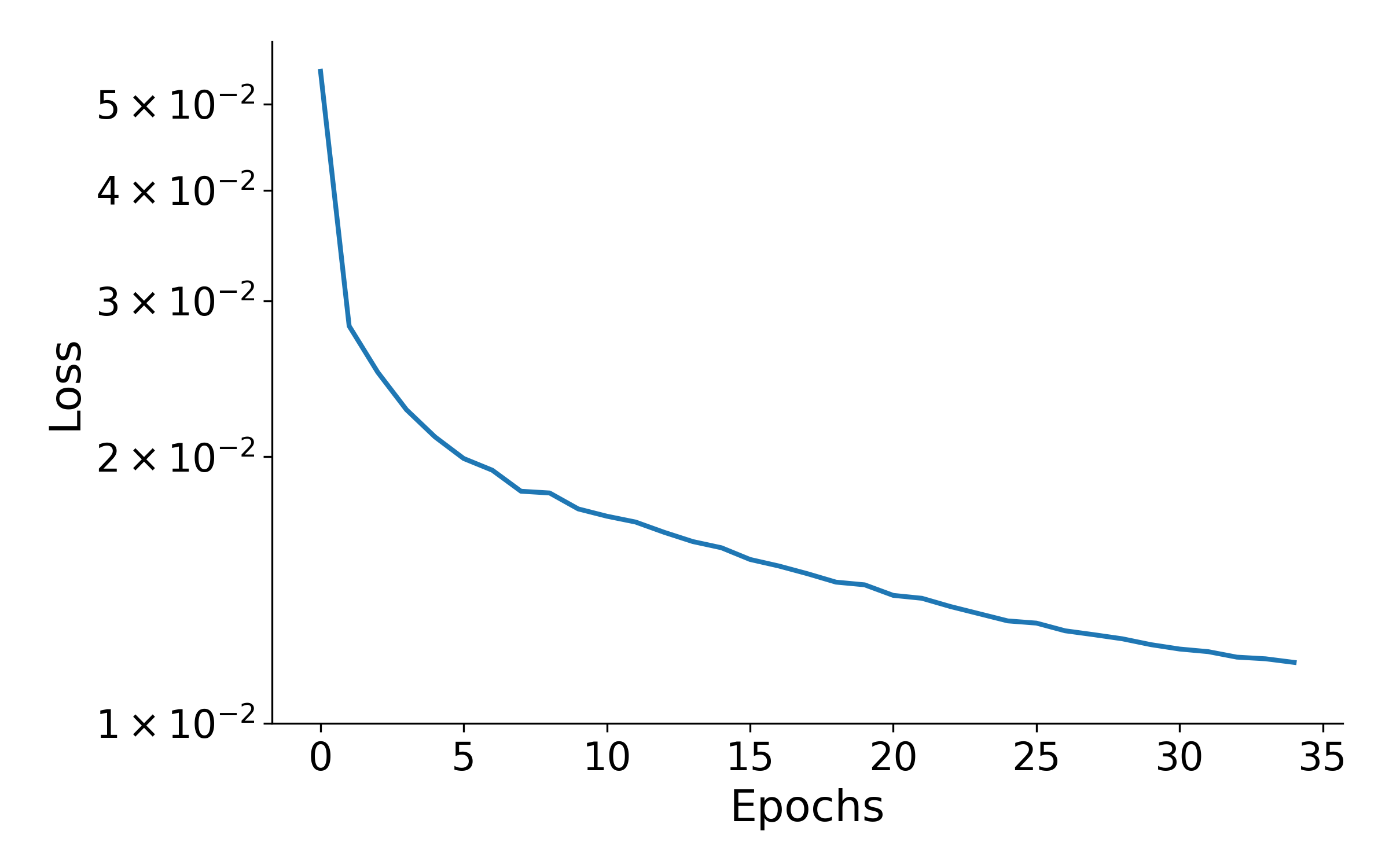}
        \caption{Bounding box loss}
        \label{fig:losssubimage2}
    \end{subfigure}
    \caption{Losses used for training gesture classification model}
    \label{fig:combined_loss}
\end{figure}

The loss values for both training and validation datasets were recorded across all epochs. \Cref{fig:combined_loss} illustrates the behavior of the two losses during training. The scalar balancing these losses was empirically determined to ensure that neither loss dominates the optimization process. Over the epochs, both loss functions showed a decreasing trend, indicating effective learning.
\vspace{-0.2in}

\subsubsection{Confusion Matrix}
\begin{figure}[t]
    \centering
    \begin{subfigure}[b]{0.44\textwidth}
        \centering
        \includegraphics[width=\textwidth]{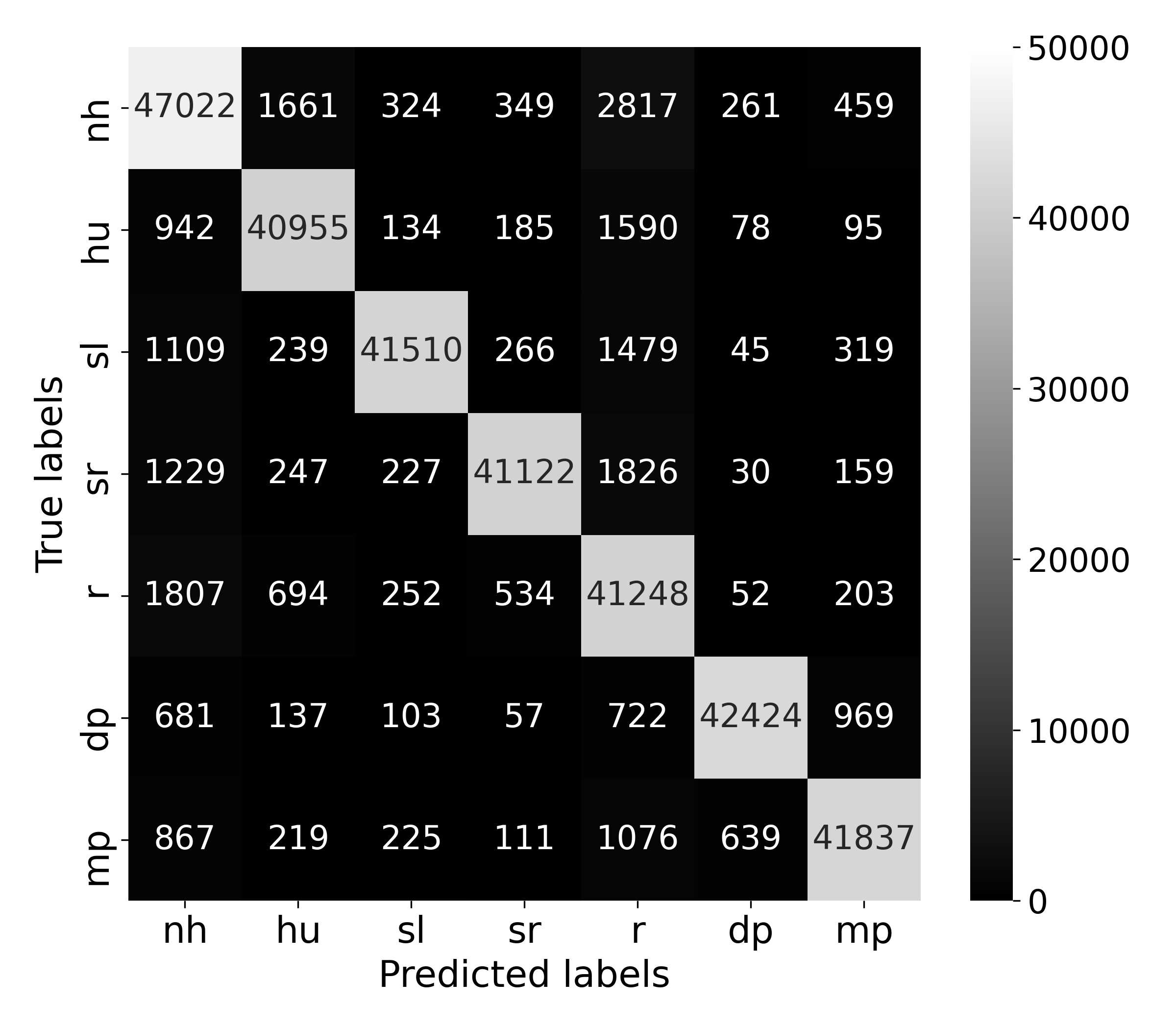}
        \caption{Confusion matrix for training data}
        \label{fig:confusionmatrix}
    \end{subfigure}
\hspace{0.05\textwidth}
    \begin{subfigure}[b]{0.44\textwidth}
        \centering
        \includegraphics[width=\textwidth]{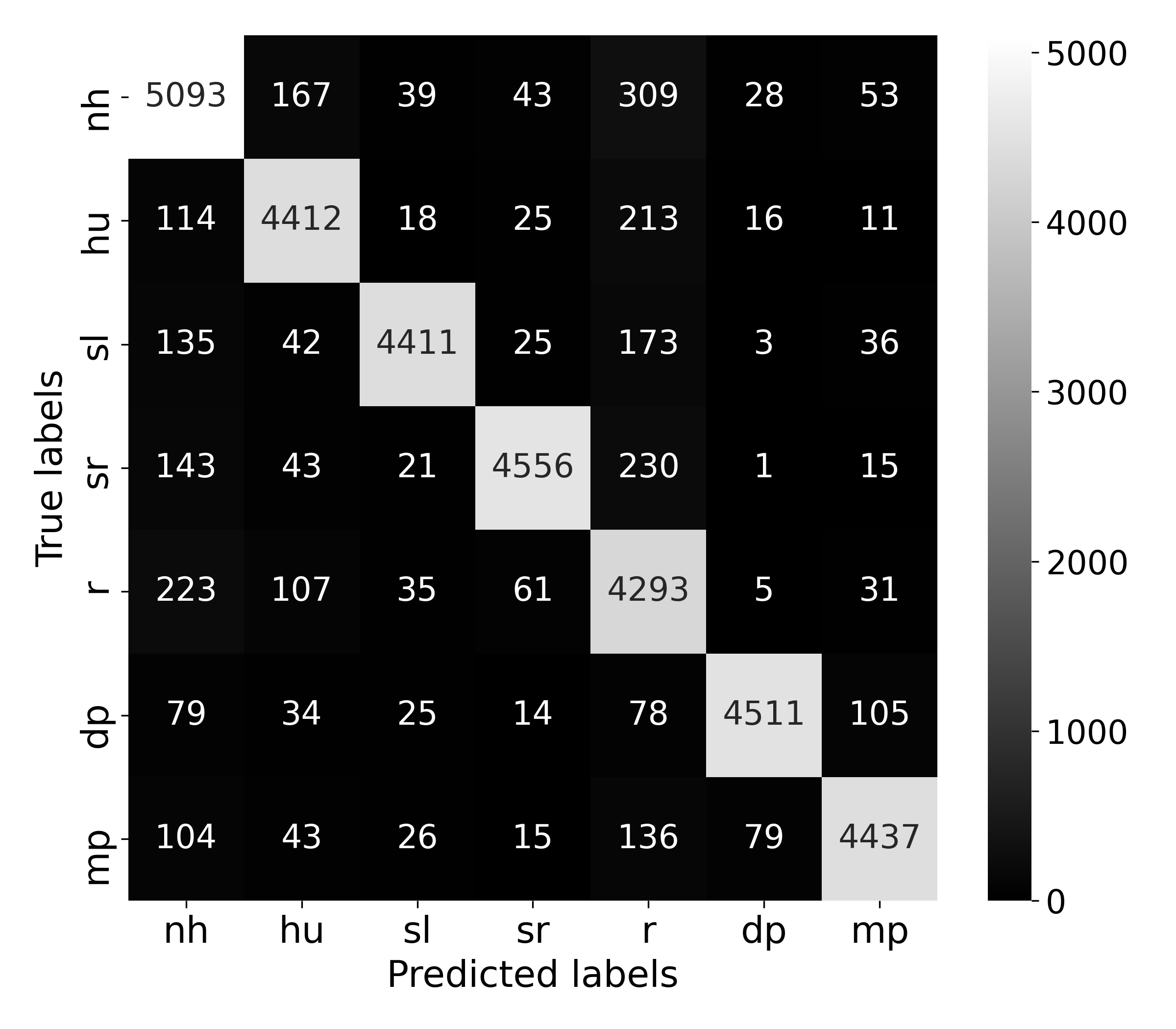}
        \caption{Confusion matrix for validation data}
        \label{fig:valconfusionmatrix}
    \end{subfigure}
    \caption{Shown here is the confusion matrix of our model
evaluated on our proposed training and validation data, where the 7 classes are detailed as: nh for no hand, hu for hand unknown, sl for swipe left, sr for swipe right, r for rest, dp for double pinch and mp for moving pinch}
    \label{fig:combined_confusion_matrix}
    \vspace{-0.2in}
\end{figure}

To further analyze the performance of our model, we computed the confusion matrix for the training and validation set. This matrix provide an insight into the classification accuracy for each class and highlights any potential issues with specific categories. \Cref{fig:combined_confusion_matrix} displays the confusion matrix for the training and validation dataset. The confusion matrix shows that the model performs well across all classes, with high true positive rates and low false positive and false negative rates. This further supports the accuracy metrics and indicates that the model has learned to distinguish between different classes effectively.
\subsection{Model Size, Power and Latency}
\subsubsection{Model Size:} The model size was evaluated by calculating the total number of parameters and the computational complexity in terms of mega floating-point operations per second (MFlops). Our proposed model has 613,836 parameters and 164 MFlops. In comparison to event based tracking methods taken from \cite{millerdurai2024eventego3d}, where the number of model parameters range from 1.3 - 11.2 million parameters and the number of Flops range from 416.8 - 3,580.0 MFlops \cite{millerdurai2024eventego3d, rudnev2021eventhands}, our model is very small.

\subsubsection{Power measurement:} The instantaneous power consumption for the model and event signal processing, measured on an NXP Nano UltraLite, was 340~mW fully utilising two of the four A-53 cores and using 30~MB of RAM. This was determined by measuring the current and voltage on the input power rail when the system was idle and running the inference. The difference minus the MIPI-USB camera controller when streaming gives the power consumption of the model. The target instantaneous power is 80~mW, this translate to an average power of approximately 16~mW derived from the number of pickups and session duration from smart watch usage, which is between 1-20\% of an 8 hour day~\cite{DeviceUsage}. To get to this the next step is to quantize the model, based on prior work we estimate that this will reduce the instantaneous power to $\sim$110~mW. The remaining reduction in power will come from architecture refinement and inference execution optimisation, for example using DSP/NPU acceleration.

\subsubsection{Latency:} The latency of our neural network running on the NXP board is measured using the high resolution clock on the CPU. The time interval is measured between the input event frame and the corresponding output detection. The averaged measured latency for the system is 60 ms, depending on factors such as temperature and compute load. In the context of HMIs, latency below 100 ms enables experience that do not require visual feedback to the user \cite{UsabilityEng} similar to capacitive touch interfaces on wearables. 

 We compare the latency of our approach to models solving similar tasks. Currently object detection models using event cameras \cite{wu2024leod, zubic2023chaos, hamaguchi2023hierarchical, li2022asynchronous, perot2020learning} report latency values in the range 9.5 - 77.2ms on either a T4 or V100 GPU. Furthermore, works on pose reconstruction \cite{rudnev2021eventhands, millerdurai2024eventego3d} report latency values in the range 2.0 - 7.1ms, again on a GPU. Our model goes a step further than only doing object detection and also performs gesture classification, while running at 60ms on a CPU. This enables the model to be run on a pair or AR glasses, which would not be feasible with prior GPU based works.
\subsection{Model Size Ablation}
To evaluate the different parameter settings in our model, we conducted an ablation study as shown in \Cref{tab:model_ablation}. This study systematically modifies number of convolutional filters and dense layer size and observes the impact on performance metrics such as training accuracy, validation accuracy, and inference speed. While increasing the parameter size does increase training and validation accuracy compared to the baseline, it also results in increased latency. The latency for Model 2 is 198ms and for Model 3 is 221ms, as compared to Model 1 latency which is 60ms. Considering the trade-off between accuracy and latency, we ultimately chose Model 1 for our purpose. 
\begin{table}[t]
    \centering
    \resizebox{\textwidth}{!}{%
    \begin{tabular}{c|c|c|c|c}
        \hline
        Model & Parameters & Training Accuracy (\%) & Validation Accuracy (\%) & Latency (ms) \\
        \hline
        Model 1 & \textbf{613,836} & 91.4 & 91.0 & \textbf{60} \\
        Model 2 & 1,916,172 & \textbf{94.4} & \textbf{93.3} & 198 \\
        Model 3 & 2,400,780 & 93.9 & 92.6 & 221 \\
        \bottomrule
    \end{tabular}
    }
    \caption{Ablation Study: Model Size vs Training and Validation Accuracy}
    \label{tab:model_ablation}
    \vspace{-0.2in}
\end{table}

\subsection{User testing}
\label{usertesting}
\subsubsection{Precision and Recall}
We evaluate the performance of our ML model with live user testing on 20 participants. For each participant, we collected the results for double-pinch, left microgesture and right microgestures. Each gesture is performed in a randomized order by a user 10 times, thereby leading to 30 gestures per user. The time between each gesture is 1.5s and the time given to complete a gesture is 2s. The precision and recall for 20 users and the learning curve for 5 naive users is given in \Cref{tab:results}. Our recall scores range from 54\% to 86\%, and precision scores range between 72\% to 92\%. The reason for double pinch scores to be lower than the swipes are that in the test app we consider the single pinch to be distinct from a double pinch, and we do not optimize the TS to be able to cleanly distinguish between them. We expect the double pinch performance to improve if we set the single pinch classification threshold to be very high. 

To start to understand how intuitive and how well the the system generalises naive users were asked to carry out the same test twice, those results are also shown in \Cref{tab:results}. This shows that both the left and right swipes show no improvement on the second attempt and results are in agreement with the original 20 users. This shows that new users can hit the same level of performance as the general population of users. This in contrast to the double pinch, which shows an increase across both recall and precision for the second attempt. This correlates with observations that the user has a smaller zone in which the pinch works well and that this can be learned through more usage. These user testing results match our experience from our recent successful demo at \href{https://www.awexr.com/usa-2024}{AWE-USA-2024}.
\vspace{-0.2cm}
\begin{table}[t]
    \centering
    \begin{tabular}{lccc}
        \toprule
        \toprule
        & Double pinch & Left swipe & Right swipe \\
        \midrule
        \midrule
        \multicolumn{4}{l}{\textbf{Recall for 20 users}} \\
         & 54$\pm$6\% & 86$\pm$4\% & 83$\pm$5\% \\
         \midrule
        \multicolumn{4}{l}{\textbf{Precision for 20 users}} \\
         & 74$\pm6$\% & 72$\pm$5\% & 92$\pm$3\% \\
        \midrule
        \midrule
        \multicolumn{4}{l}{\textbf{Learning curve for 5 naive users}} \\
        \multicolumn{4}{l}{\textbf{Recall}} \\
         & 49$\pm$6\% $\rightarrow$ 66$\pm$6\% & 87$\pm$6\% $\rightarrow$ 85$\pm$3\% & 76$\pm$14\% $\rightarrow$ 82$\pm$4\% \\
         \midrule
        \multicolumn{4}{l}{\textbf{Precision}} \\
         & 67$\pm$15\% $\rightarrow$ 89$\pm$14\% & 81$\pm$7\% $\rightarrow$ 73$\pm$3\% & 91$\pm$5\% $\rightarrow$ 93$\pm$8\% \\
        \bottomrule
        \vspace{-0.1in}
    \end{tabular}
    \caption{Recall and Precision for 20 users and Learning Curve for 5 naive users}
    \label{tab:results}
    \vspace{-0.2in}
\end{table}

\section{Conclusion}
\label{sec:conclusion}
\vspace{-0.2cm}
In this work we demonstrated a novel HMI event-based vision system for smart eyewear, optimised for state of the art natural hand interactions. This system has several advantages over existing solutions to the eyewear HMI problem. These include human centric ergonomic design and discreetness afforded by the smart glasses form factor, enabled by the compactness of the sensor hardware, and a low-power machine learning model architecture capable of detecting subtle gestures (with the potential for further power-performance optimisation). The use of event sensors keeps the power within what is deliverable by a battery powered device, and enables future work to generalise its performance to most indoor and outdoor lighting conditions. Our ML model's efficiency also implies the headroom to extend the number of different gestures which are detected and classified.




%
%
\bibliographystyle{splncs04}
\bibliography{main}

\clearpage
\appendix
\section{Appendix}

\subsection{Testing App}
\begin{figure}[htbp]
    \centering
    \includegraphics[width=0.3\textwidth]{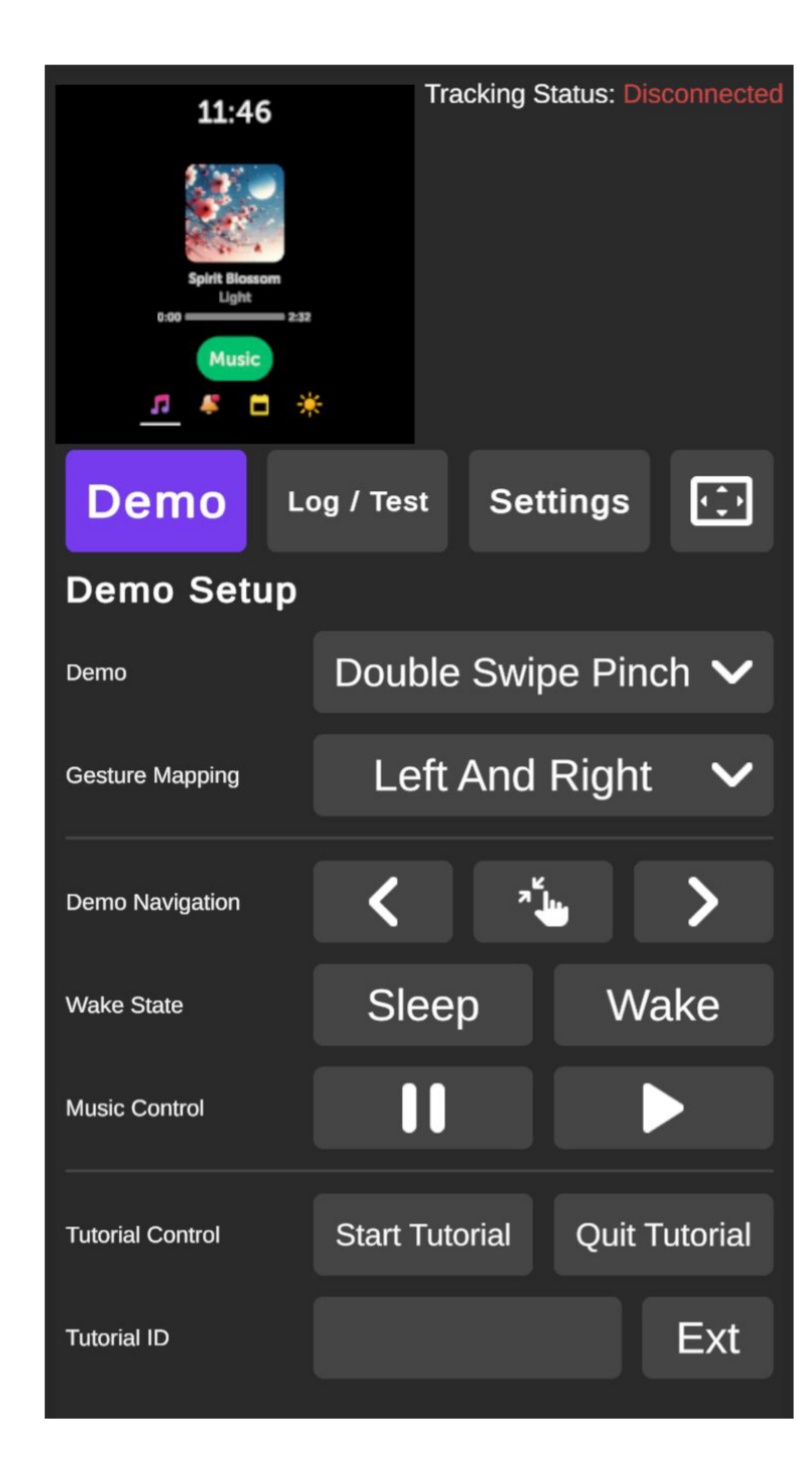}
    \caption{Our custom testing app to measure user precision and recall}
    \label{fig:customapp}
\vspace{-0.2in}
\end{figure}

We developed a custom user testing app, shown in \Cref{fig:customapp}, that provides a user with a randomised set of gestures, with a configurable amount of repetitions. These gestures are presented to the user where they must be actioned within a specific time frame. Each user had one attempt to make the correct gesture, otherwise it would be classified as a failure (the wrong gesture) or a timeout (no gesture received). These statistics informed us of the precision of the gestures, as well as the recall. An example of the user wearing the smart glasses with the event sensor and controlling media is given in \Cref{fig:finalsetup}. The system is very responsive, immediately detecting the gesture in most cases

\label{sec:appendix}


\end{document}